\documentclass[11pt]{article}
\usepackage{booktabs}

\usepackage[preprint]{acl}
\usepackage{amsmath}

\usepackage{times}
\usepackage{latexsym}
\usepackage{multirow}
\usepackage[T1]{fontenc}
\usepackage{listings}
\usepackage{caption}
\usepackage[utf8]{inputenc}
\usepackage{microtype}
\usepackage{makecell}
\usepackage{inconsolata}

\usepackage{graphicx}
\newcommand{\eg}{\textit{e.g.,}\ }
\newcommand{\ie}{\textit{i.e.,}\ }

\hypersetup{
    hidelinks,
    breaklinks=true
}
\usepackage[all]{hypcap}
\usepackage{comment}

\title{SocraticKG: Knowledge Graph Construction\\via QA-Driven Fact Extraction}

\author{
  \textbf{Sanghyeok Choi}\textsuperscript{1}\thanks{\ \ Equal contribution} \quad
  \textbf{Woosang Jeon}\textsuperscript{1,2}\footnotemark[1] \quad
  \textbf{Kyuseok Yang}\textsuperscript{1} \quad
  \textbf{Taehyeong Kim}\textsuperscript{1,2,3,4}\thanks{\ \ Corresponding author} \\
  \\ 
  \textsuperscript{1}Department of Biosystems Engineering, Seoul National University \\
  \textsuperscript{2}Artificial Intelligence Institute, Seoul National University \\
  \textsuperscript{3}Interdisciplinary Program in Artificial Intelligence, Seoul National University \\
  \textsuperscript{4}Interdisciplinary Program in Cognitive Science, Seoul National University \\
  \texttt{\{cholsang83, jwoosang1, kyuseok0603, taehyeong.kim\}@snu.ac.kr}
}

\begin{document}
\maketitle
\begin{abstract}
Constructing Knowledge Graphs (KGs) from unstructured text provides a structured framework for knowledge representation and reasoning, yet current LLM-based approaches struggle with a fundamental trade-off: factual coverage often leads to relational fragmentation, while premature consolidation causes information loss.
To address this, we propose SocraticKG, an automated KG construction method that introduces question-answer pairs as a structured intermediate representation to systematically unfold document-level semantics prior to triple extraction.
By employing 5W1H-guided QA expansion, SocraticKG captures contextual dependencies and implicit relational links typically lost in direct KG extraction pipelines, providing explicit grounding in the source document that helps mitigate implicit reasoning errors.
Evaluation on the MINE benchmark and HotpotQA downstream task demonstrates that our approach effectively addresses the coverage-connectivity trade-off, achieving superior factual retention and structural cohesion while supporting complex multi-hop reasoning.

\end{abstract}

\section{Introduction}

\begin{figure*}[t]
    \phantomsection
    \centering
    \includegraphics[width=0.99\textwidth]{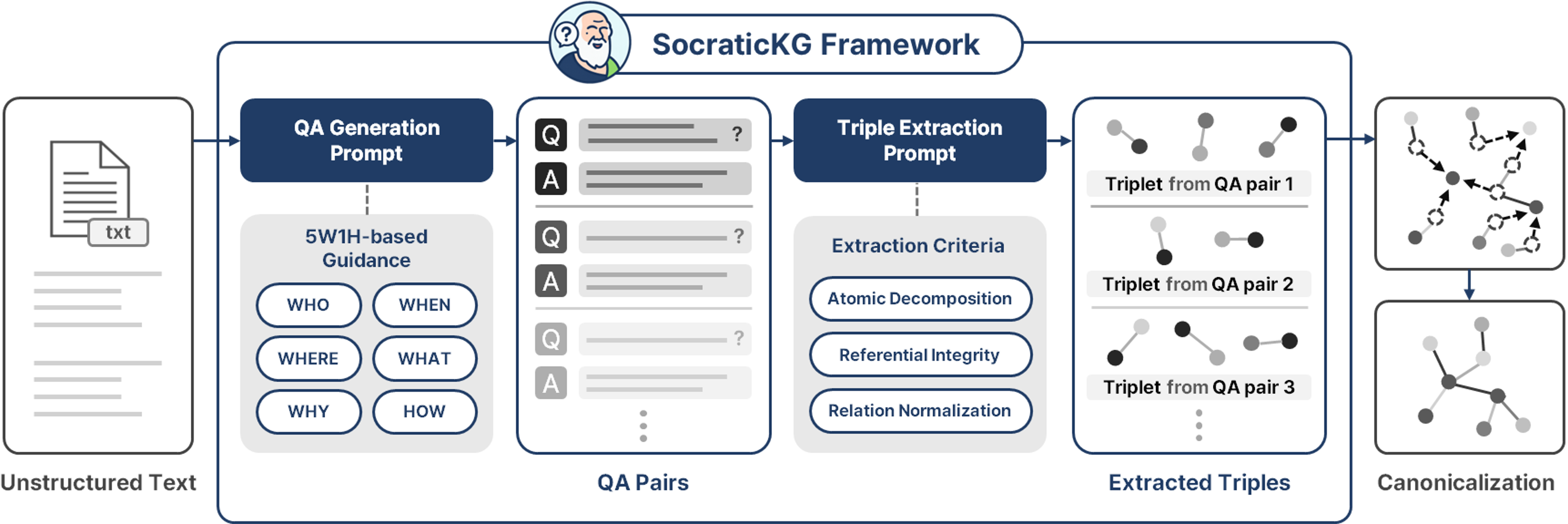}
    \caption{The overall architecture of the SocraticKG framework. Given unstructured text, the method first generates atomic QA pairs through 5W1H-guided questioning, then extracts triples from these QA pairs, and finally canonicalizes the triples to produce a cohesive knowledge graph.}
    \label{fig:main}
\end{figure*}

As large language models (LLMs) are widely used in knowledge-intensive applications, concerns surrounding factual reliability, interpretability, and grounding have become more pronounced \citep{ji2023survey, huang2025survey}.
While Retrieval-Augmented Generation (RAG) addresses these concerns by anchoring models to external sources, it often struggles with fragmented contexts and shallow integration of complex facts \citep{lewis2020retrieval, gao2023retrieval}.
In response, Knowledge Graphs (KGs) have re-emerged as a complementary solution, providing a structured and verifiable backbone for explicit knowledge representation and reasoning \citep{pan2023large, Rajabiexplainable2024}.
However, the reliance on manual curation has historically limited the availability of domain-specific KGs, thereby motivating growing interest in automated construction methods that can scale to diverse and large-scale text sources \citep{ren2024survey}.

Recent advances in LLMs have enabled more semantically grounded approaches to knowledge graph construction, moving beyond rule-based pattern matching toward methods that leverage neural reasoning to interpret unstructured text \citep{zhu2024llms}.
Current approaches address the construction challenge through different strategies.
Some methods focus on capturing explicit factual mentions in a single pass, extracting triples directly from text \citep{cabot2021rebel, shang2022onerel, zhang-soh-2024-extract}.
Others adopt consolidation-centric strategies, organizing extracted facts around pre-identified entity structures to improve graph coherence \citep{zhong2021frustratingly, ye2022packed, wei2023chatie, mo2025kggen}.

However, these approaches face a persistent challenge: fully externalizing the narrative logic of source documents into structured graphs.
The resulting knowledge graphs often struggle with a fundamental tension between factual coverage and structural coherence.
Graphs may contain many facts but remain fragmented with weak semantic connectivity, or they may be well-organized yet incomplete, having filtered out contextual nuances that do not conform to predefined structures.
At the core of this challenge lies the difficulty of balancing comprehensive information extraction with meaningful connectivity across the graph.

To address this limitation, we draw inspiration from how humans naturally process and organize information from text.
Rather than attempting to extract structured knowledge in a single step, human comprehension is fundamentally interrogative: readers construct understanding by progressively clarifying salient concepts through active inquiry \citep{graesser1994question, ambrose2010learning}.
This process of interrogative learning serves as a natural scaffold for organizing complex information.
Question-Answering (QA), in particular, facilitates focused attention and explicit articulation of relationships that might otherwise remain implicit in direct extraction \citep{wu2020corefqa}.

Building on this insight, we propose SocraticKG (SoKG), a method that treats QA not merely as a retrieval mechanism, but as a structured intermediate representation that systematically unfolds document-level semantics prior to graph construction \citep{fitzgerald2018large, cohen2023qa}.
SoKG employs a structured interrogative framework based on the 5W1H framework (\textit{who}, \textit{what}, \textit{when}, \textit{where}, \textit{why}, and \textit{how}) to generate document-grounded QA pairs that capture key concepts, relationships, and contextual dependencies.
This QA-mediated expansion articulates implicit connections and contextual nuances in explicit natural language format.
The resulting intermediate representation facilitates more consistent and complete triple extraction by providing well-defined semantic units rather than requiring simultaneous resolution of semantics and structure.
These extracted triples are then unified through a canonicalization process \citep{mo2025kggen} that resolves surface-form variations and consolidates the graph into a coherent structure.

We evaluate our proposed method on the MINE (Measure of Information in Nodes and Edges) benchmark \citep{mo2025kggen} and HotpotQA~\citep{yang2018hotpotqa} as a downstream multi-hop reasoning task.
Our results demonstrate that SocraticKG consistently outperforms state-of-the-art counterparts across multiple LLM backbones, achieving superior factual retention while producing more densely connected and less fragmented graphs, yielding further gains on complex multi-hop reasoning.\footnote{Our code is publicly available at \url{https://github.com/LABA-SNU/SocraticKG}.}

In summary, we make the following contributions in this work:
\begin{itemize}
    \item We propose SocraticKG, a QA-mediated method for knowledge graph construction that formalizes question-answering as a semantic scaffold for unfolding document narratives and explicitly articulating implicit connections prior to structural extraction.
    \item We introduce 5W1H-guided QA expansion as a systematic approach for surfacing latent dependencies typically overlooked in direct extraction, thereby improving factual coverage while reducing implicit reasoning errors.
    \item We demonstrate that our approach mitigates structural fragmentation and information loss, achieving superior factual retention and recoverability across various LLMs.
\end{itemize}

\section{Related Work}
\subsection{Direct Triple Extraction}
Knowledge Graph (KG) construction has evolved from conventional Open Information Extraction (OpenIE) \citep{etzioni2008open, fader2011identifying} to modern approaches that extract triples directly via LLMs \citep{cabot2021rebel, bi2024codekgc, zhang-soh-2024-extract}.
While OpenIE is constrained by surface linguistic patterns \citep{niklaus2018survey}, such direct extraction methods leverage LLM reasoning capabilities to bridge semantic gaps without explicit intermediate representations.

However, this direct extraction approach often limits the model to capturing surface-level, explicit mentions while overlooking the latent logical ties that bind them.
As noted by \citet{zhu2024llms, meher2025link}, this approach often yields shallow factual coverage, often producing fragmented subgraphs that lack the connectivity required for effective graph-based reasoning.

\subsection{Consolidation-Centric Strategies}
To address the fragmentation issues, various pipelines emphasize structural coherence through post-extraction consolidation.
These entity-first approaches organize extracted facts by first identifying key entities, then structuring relations around this pre-established entity framework.
GraphRAG \citep{edge2024local} builds a global index of entities and relationships partitioned into hierarchical communities for query-focused summarization, whereas KGGen \citep{mo2025kggen} emphasizes clustering-based canonicalization of entities and relations to produce compact and reusable knowledge graphs.
Similarly, CLARE \citep{henry2025clare} anchors its relational extraction on initial entity identification to ensure semantic precision within consolidated text.

Despite their effectiveness in organizing triples, these consolidation-focused strategies can act as a representational bottleneck \citep{ye2022generative}.
When entity sets are fixed early in the pipeline, relations or contextual dependencies that do not conform to the initial entity structure may be excluded.
This sequencing effectively prioritizes structural utility over factual density, potentially under-representing the document’s latent relations.

\subsection{Transform-Then-Extract Approaches}
To reduce extraction complexity, various approaches employ a two-stage process: first transforming raw text through intermediate representations, then extracting triples from the transformed output.
Common transformation strategies include coreference resolution to handle referential expressions \citep{manning2014stanford, cetto2018graphene} and syntactic sentence decomposition to simplify complex structures \citep{niklaus2019transforming, niklaus2022complex}.
CoDe-KG \citep{Anuyah_2025}, for instance, leverages human-guided prompt intervention to incorporate these transformation tasks, ensuring structural clarity prior to extraction.

These transformation-based approaches operate primarily at the sentence level, focusing on local syntactic normalization rather than document-level semantic organization.
While effective for resolving surface-level ambiguities within individual sentences, they do not systematically capture cross-sentence dependencies or contextual relationships that span the document.
This limits their ability to externalize the broader narrative structure and global semantics required for comprehensive knowledge graph construction.

\subsection{QA for Knowledge Extraction}
Question-answering has been widely used to elicit structured information from text \citep{levy2017zero, li2019entity, du2020event}, by leveraging the cognitive process of interrogative inquiry, which facilitates the construction of situation models \citep{chi1989self, graesser1994question}.
Recent extraction methods, such as StoryNet \citep{nagireddy2021storynet} and ChatIE \citep{wei2023chatie}, incorporate QA-driven prompting as a core component of their extraction pipelines.

However, these approaches treat QA pairs as transient artifacts, generating and consuming them within a single extraction pass, without formalizing them as an intermediate representation for organizing document-level semantics.
As a result, they lack systematic question generation and struggle to surface implicit relational and contextual dependencies prior to triple extraction.

While recent work has explored QA as an intermediate step for interpretable knowledge construction \citep{aneja2025interpretable}, it primarily emphasizes retrieval utility through factual restatement, rather than semantic organization.
Collectively, these gaps suggest that formalizing QA as a structured intermediate representation provides a more robust foundation for construction, particularly when systematic inquiry is used to proactively externalize latent relational and causal dependencies.

\section{Methods}
SoKG introduces QA pairs as a structured intermediate representation for LLM-based KG construction.
Rather than prompting LLMs to extract triples directly from raw text, our approach first decomposes the document into explicit QA pairs that resolve contextual dependencies and referential ambiguities in natural language.
These QA pairs are then mapped to atomic triples and unified through canonicalization to produce the final KG.

\subsection{5W1H-Guided QA Generation}
This stage transforms document into a collection of discrete, self-contained QA pairs.
To ensure comprehensive coverage of the factual content in the text, we design a prompt strategy based on two core principles: systematic questioning and contextual independence (detailed prompt in Appendix~\ref{sec:5w1h_prompts}).

\paragraph{Detailed Questioning via 5W1H}
We leverage the 5W1H framework to guide systematic question formulation.
The LLM generates multiple questions spanning all six categories and diverse aspects of the document.
As a result, the resulting QA pairs capture both surface-level entities and complex dependencies, including causal rationales (\textit{why}) and procedural details (\textit{how}).

\paragraph{Contextual Independence}
To ensure each QA pair functions as a standalone unit, we instruct the LLM to generate answers that are fully understandable without referencing the original source text.
Specifically, the model is required to replace pronouns (\eg \textit{it}, \textit{they}) with their explicit entity names, resolving referential ambiguities.
This constraint prevents information loss when each QA pair is processed individually in the triple extraction phase.

\subsection{Triple Extraction from QA}
This stage transforms the QA pairs into structured triples by treating each pair as an independent extraction unit.
Operating on these logically self-contained units allows the extraction process to focus on well-defined semantic boundaries, reducing errors common in direct extraction from long, complex texts.
To achieve this, the LLM is instructed to follow three specific constraints (detailed prompt in Appendix~\ref{sec:triple_prompts}).

\paragraph{Atomic Decomposition}
The model decomposes each QA pair into separate, atomic triples, capturing fine-grained facts from both the inquiry and the response to maximize factual richness.

\paragraph{Entity Clarity}
All entities are expressed as specific noun phrases, and any triple containing ambiguous pronouns is discarded. 
This ensures that every extracted fact is self-contained and grounded in clear evidence.

\paragraph{Simplified Relations}
Predicates are distilled into concise verb phrases to reduce surface-form variations, facilitating subsequent canonicalization.
The model is instructed to skip extraction if the relatio nship remains ambiguous.

\subsection{Graph Construction from Triples}
\label{subsec:graph-construction-triples}
The final stage unifies discrete triples into a cohesive graph structure.
Since extraction occurs across independent QA units, the raw set often contains redundant or synonymous mentions for the same concept.
To resolve these redundancies, we adopt the canonicalization procedure from~\citet{mo2025kggen}, which combines embedding-based clustering with LLM-based refinement.

The canonicalization process is performed independently on entities and relations through a cluster-then-refine process.
First, semantic embeddings are generated for all unique entities and relations using a text embedding model.
To narrow the search space, these embeddings are partitioned into clusters of a manageable size for entities and relations respectively via K-means clustering.
Within each cluster, the top-$k$ potential matches for each anchor are identified by balancing dense semantic similarity with sparse lexical overlap (BM25).
Finally, synonyms and abbreviations are resolved by an LLM, which maps these variants to a single representative form to consolidate fragmented triples into a cohesive, canonicalized graph.   

\begin{figure*}[t]
    \centering
    \includegraphics[width=0.99\textwidth]{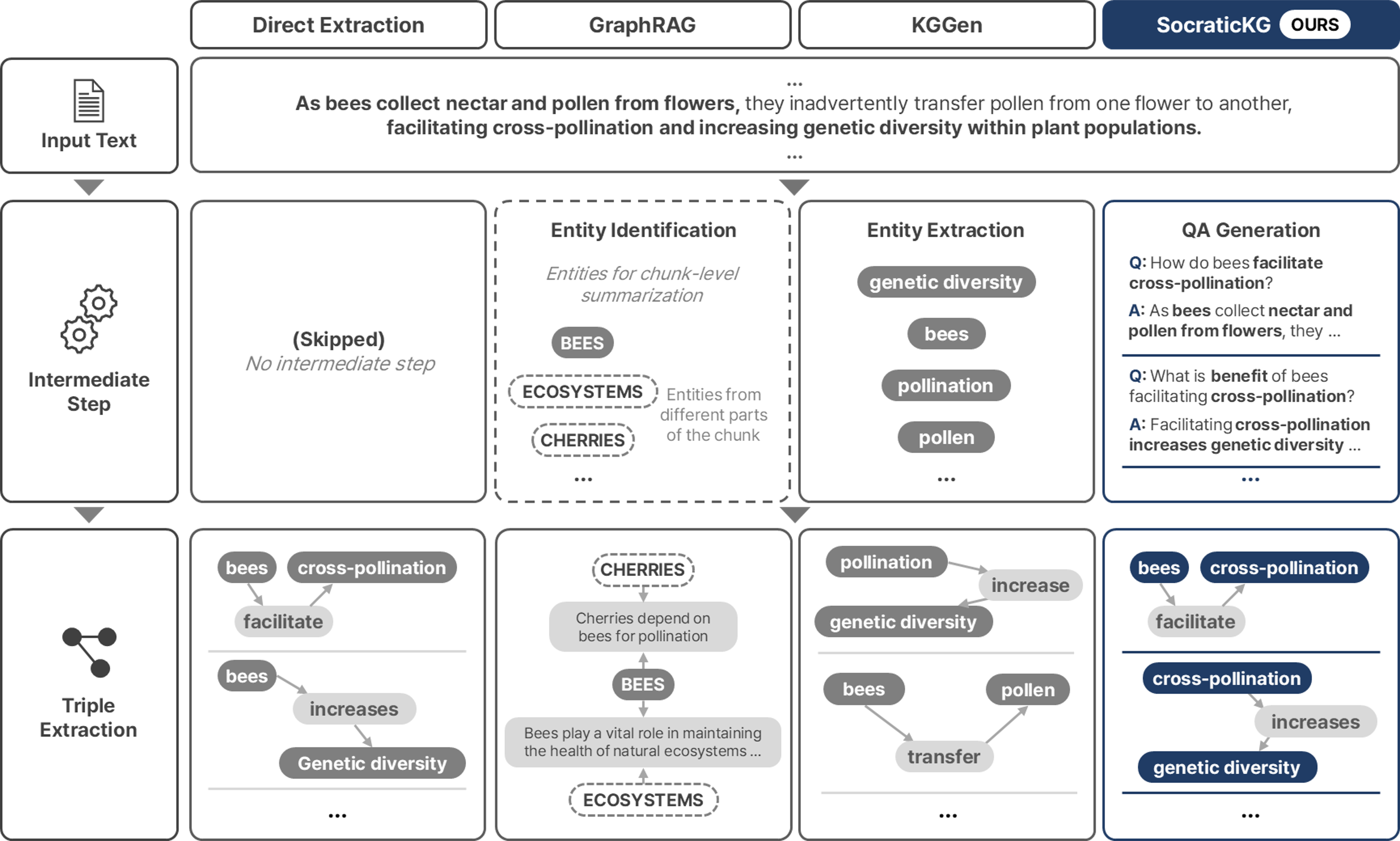}
    \caption{Comparison of extraction pipelines using an example output from Gemini-2.5-flash-lite. While baseline pipelines often miss the syntactic connection in complex sentences, failing to recover the causal link between bees and genetic diversity, SoKG leverages QA-driven reasoning to explicitly reconstruct the intermediate concept. As a result, SoKG successfully recovers the complete causal chain (bees $\rightarrow$ cross-pollination $\rightarrow$ genetic diversity), whereas baselines tend to simplify or fragment this relationship.}
    \label{fig:case_study}
\end{figure*}

\section{Experiments}

We evaluate SoKG from two complementary perspectives:
source-information preservation and downstream reasoning utility.
For the former, we utilize the MINE benchmark~\citep{mo2025kggen}, designed to quantify the information gap between raw text and its graph representation by measuring how much source information is recoverable.
The benchmark comprises 100 diverse articles, each paired with 15 verified atomic facts, providing a rigorous evaluation framework across 1,500 independent factual instances.
Following the benchmark protocol, we constructed one KG per article and evaluated each graph in terms of factual retention and structural characteristics.

For downstream reasoning utility, we examine whether SoKG's structural advantages translate into practical gains beyond source-information preservation.
To this end, we conducted experiments on HotpotQA~\citep{yang2018hotpotqa} using 800 ``Hard'' Bridge samples.
These samples require identifying an intermediate entity that connects disparate pieces of evidence, making them well suited for evaluating graph-based multi-hop reasoning.

\subsection{Evaluation Metrics}
\paragraph{Factual Retention Score}
As the primary metric, we measured the proportion of ground-truth facts successfully recovered from the constructed KGs.
Following the MINE benchmark protocol, we retrieved a local subgraph for each fact, consisting of the top-8 nodes most semantically similar to the target statement and their 2-hop neighbors.
An LLM-judge then determined whether the fact was logically supported by the retrieved subgraph context.
The score represents the percentage of verifiable facts, reflecting how well the graph preserves information from the source text for downstream tasks such as retrieval and reasoning.

\begin{table*}[t]
  \centering
  \small
  \setlength{\tabcolsep}{0pt}
  \renewcommand{\arraystretch}{1.1}
  \begin{tabular*}{\textwidth}{@{\extracolsep{\fill}} l ccccc}
    \toprule
    \textbf{Method} & \textbf{Qwen-2.5} & \textbf{GPT-4o-mini} & \textbf{GPT-4o} & \textbf{Gemini-2.5} & \textbf{Claude-4} \\
    \midrule
    Direct Extraction & 66.5 & 68.5 & 78.1 & 84.6 & 86.8 \\
    GraphRAG          & 59.7 & 49.5 & 49.3 & 48.5 & 52.3 \\
    KGGen             & 56.7 & 44.3 & 66.4 & 62.5 & 69.1 \\
    \midrule
    SoKG (w/o 5W1H)   & 67.1 & 80.5 & 83.5 & 85.6 & 94.6 \\
    \textbf{SoKG (Ours)} & \textbf{73.4} & \textbf{83.9} & \textbf{89.3} & \textbf{87.7} & \textbf{96.3} \\
    \bottomrule
  \end{tabular*}
  \caption{Comparison of factual retention scores (\%) on the MINE benchmark. SoKG consistently achieves the highest performance across all evaluated models.
  The vanilla variant (\ie SoKG w/o 5W1H) shows how the 5W1H scaffold captures procedural and causal facts to improve factual consistency even on smaller models like Qwen-2.5.}
  \label{tab:main_results}
\end{table*}

\paragraph{Structural Cohesion and Density}
To analyze the organization and coherence of the KGs, we investigated:
\begin{itemize}
    \item \textbf{Average Degree (Deg):}
    The average number of unique neighboring nodes per node, capturing the local connectivity density of the graph~\citep{barabasi2013network}.
    It reflects how many distinct entities a node is connected to, irrespective of relation direction.
    Following standard practice for undirected graphs, we compute the average degree as
    \[\text{Deg} = \frac{2E}{N},\]
    where $N$ denotes the number of nodes and $E$ the number of edges.

    \item \textbf{Triple Count (\#Tri):} The total number of atomic facts externalized in the graph.
    \item \textbf{Normalized Fragmentation Index (NFI):}
    Motivated by the notion of graph fragmentation as the decomposition of a network into disconnected components~\citep{borgatti2003key}, we define a component-based metric as:
    \[ \text{NFI} = \frac{C - 1}{N - 1}, \]
    where $C$ denotes the number of connected components and $N$ is the total number of nodes ($N \ge 2$).
    This formulation normalizes fragmentation to the unit interval $[0,1]$, where 0 corresponds to a fully connected graph ($C=1$) and 1 indicates a completely fragmented network ($C=N$).
\end{itemize}

\paragraph{Multi-hop Reasoning Accuracy}
For downstream evaluation on HotpotQA, we measured answer accuracy on the Hard Bridge samples.
For each question, we constructed a KG from the associated source documents, retrieved a top-1 seed triple via embedding similarity, and expanded its neighborhood to 2-hop and 3-hop depths. 
An answer was marked correct when an LLM answering model, given only the retrieved subgraph context, produced a response matching the gold answer.

\subsection{Comparative Analysis Design}
\paragraph{KG Construction Methods}
We compared SoKG against three representative approaches in the current landscape of LLM-based KG construction.
To ensure a valid comparison, we selected the comparative methods that operated in autonomous and open-domain settings without pre-defined schemas or human intervention.
Figure~\ref{fig:case_study} summarizes the procedures of these methods.

\begin{itemize}
    \item \textbf{Direct Extraction:}
    A single-pass extraction strategy where triples are generated directly from raw text (Appendix~\ref{sec:direct_extraction_prompt}).
    For fair comparison, we apply the identical canonicalization procedure used in KGGen and SoKG to consolidate the extracted triples.
    It serves as a primary benchmark for the LLM’s implicit reasoning capability without the benefit of intermediate semantic scaffolding.
    \item \textbf{GraphRAG:}
    A prominent solution across industry and academia for global, query-focused entity indexing.
    We utilize Microsoft's official implementation for hierarchical community detection and aggregation, providing a benchmark against the widely adopted text-summary-based method.
    \item \textbf{KGGen:}
    A recent state-of-the-art method focusing on entity-centric extraction and structural consolidation.
    It serves as a primary comparative method for evaluating factual retention and structural cohesion in open-domain.
    \item \textbf{SoKG (w/o 5W1H):} 
    An ablated variant of SoKG that retains QA pairs as its intermediate representation but replaces the 5W1H-guided inquiry with generic QA.
    This design isolates the contribution of the 5W1H-guided scaffold to evaluate its impact.
    \item \textbf{SoKG (Ours):}
    Our proposed approach utilizing 5W1H-guided QA generation to systematically construct KGs from source documents.
    Unless otherwise specified, SoKG refers to this complete implementation.
\end{itemize}

\paragraph{Evaluation across LLMs}
To assess robustness across varying LLM architectures and scales, we evaluated the selected KG construction methods on five LLMs: GPT-4o, GPT-4o-mini, Gemini-2.5 (Gemini-2.5-Flash-Lite), Qwen-2.5 (Qwen2.5-7B-Instruct), and Claude-4 (Claude-4-Sonnet).

\paragraph{Downstream Baseline Comparison}
For the HotpotQA experiments, we include a Naive RAG baseline that retrieves flat text chunks via embedding similarity.
Chunk sizes were matched to the average character length of graph-based retrieval contexts produced by SoKG for each backbone, ensuring both methods operate on comparable context lengths: 100 characters $\times$ Top-3 for Qwen-2.5 and 500 characters $\times$ Top-3 for Claude-4.

\begin{table*}[t]
  \centering
  \small
  \resizebox{\textwidth}{!}{%
  \begin{tabular}{l ccc ccc ccc ccc ccc}
    \toprule
    \multirow{2.3}{*}{\textbf{Method}}
      & \multicolumn{3}{c}{\textbf{Qwen-2.5}}
      & \multicolumn{3}{c}{\textbf{GPT-4o-mini}}
      & \multicolumn{3}{c}{\textbf{GPT-4o}}
      & \multicolumn{3}{c}{\textbf{Gemini-2.5}}
      & \multicolumn{3}{c}{\textbf{Claude-4}} \\
    \cmidrule(lr){2-4}\cmidrule(lr){5-7}\cmidrule(lr){8-10}
    \cmidrule(lr){11-13}\cmidrule(lr){14-16}
      & \textbf{N} & \textbf{E} & \textbf{Deg}
      & \textbf{N} & \textbf{E} & \textbf{Deg}
      & \textbf{N} & \textbf{E} & \textbf{Deg}
      & \textbf{N} & \textbf{E} & \textbf{Deg}
      & \textbf{N} & \textbf{E} & \textbf{Deg} \\
    \midrule
    Direct Extraction
      & 21.7 & 17.3 & 1.60
      & 33.5 & 28.1 & 1.69
      & 33.9 & 27.4 & 1.62
      & 58.4 & 64.1 & 2.20
      & 46.4 & 40.8 & 1.77 \\
    GraphRAG
      & 19.8 & 19.0 & \textbf{2.00}
      & 11.2 & 10.2 & 1.84
      & 11.3 & 9.70 & 1.75
      & 15.4 & 17.7 & 2.35
      & 14.6 & 16.2 & 2.20 \\
    KGGen
      & 28.1 & 22.1 & 1.56
      & 19.3 & 16.7 & 1.75
      & 33.2 & 28.9 & 1.74
      & 38.1 & 43.2 & 2.23
      & 57.2 & 58.9 & 2.07 \\

    \midrule
    SoKG (w/o 5W1H)
      & 28.0 & 25.4 & 1.81
      & 49.2 & 50.5 & 2.06
      & 51.9 & 49.1 & 1.89
      & 58.0 & 67.8 & 2.34
      & 84.2 & 94.5 & 2.25 \\
    
    \textbf{SoKG (Ours)}
      & \textbf{34.9} & \textbf{34.1} & 1.96
      & \textbf{57.9} & \textbf{62.2} & \textbf{2.16}
      & \textbf{62.3} & \textbf{60.5} & \textbf{1.95}
      & \textbf{65.7} & \textbf{80.8} & \textbf{2.47}
      & \textbf{104.2} & \textbf{128.4} & \textbf{2.48} \\
    \bottomrule
  \end{tabular}}
  \caption{Topological characteristics averaged over the 100 articles in the MINE benchmark. N, E, and Deg denote the mean count of Nodes, Edges, and Average Degree per graph, respectively.
  SoKG consistently expands the knowledge scale while maintaining high connectivity density across all backbones.}
  \label{tab:statistics}
\end{table*}

\begin{table*}[t]
  \centering
  \small
  \setlength{\tabcolsep}{0pt}
  \renewcommand{\arraystretch}{1.1}
  \begin{tabular*}{\textwidth}{@{\extracolsep{\fill}} l cc cc cc cc cc}
    \toprule
    \multirow{2.3}{*}{\textbf{Method}} 
      & \multicolumn{2}{c}{\textbf{Qwen-2.5}} 
      & \multicolumn{2}{c}{\textbf{GPT-4o-mini}} 
      & \multicolumn{2}{c}{\textbf{GPT-4o}} 
      & \multicolumn{2}{c}{\textbf{Gemini-2.5}} 
      & \multicolumn{2}{c}{\textbf{Claude-4}} \\
    \cmidrule(lr){2-3} \cmidrule(lr){4-5} \cmidrule(lr){6-7} \cmidrule(lr){8-9} \cmidrule(lr){10-11}
      & \textbf{NFI} & \textbf{\#Tri}
      & \textbf{NFI} & \textbf{\#Tri}
      & \textbf{NFI} & \textbf{\#Tri}
      & \textbf{NFI} & \textbf{\#Tri}
      & \textbf{NFI} & \textbf{\#Tri} \\
    \midrule
    Direct Extraction & 0.162 & 1,955 & 0.145 & 3,100 & 0.172 & 2,941 & 0.038 & 7,315 & 0.127 & 4,417 \\
    GraphRAG          & 0.084 & 1,981 & \textbf{0.038} & 1,076 & \textbf{0.083} & 1,009 & 0.036 & 1,848 & 0.067 & 1,590 \\
    KGGen             & 0.187 & 2,375 & 0.091 & 1,942 & 0.112 & 3,089 & 0.030 & 5,301 & 0.052 & 6,391 \\
    \midrule
    SoKG (w/o 5W1H)   & 0.106 & 2,871 & 0.059 & 5,646 & 0.092 & 5,345 & 0.034 & 7,875 & 0.056 & 10,511\\
    \textbf{SoKG (Ours)} & \textbf{0.078} & \textbf{3,958} & 0.047 & \textbf{7,069} & 0.086 & \textbf{6,627} & \textbf{0.023} & \textbf{9,612} & \textbf{0.039} & \textbf{14,849} \\
    \bottomrule
  \end{tabular*}
  \caption{Comparison of graph fragmentation averaged over the 100 articles (NFI; lower is better) and total extracted information volume summed over the 100 articles (\#Tri). The results demonstrate that SoKG effectively resolves the trade-off between knowledge coverage and structural connectivity, maintaining high graph cohesion even as the volume of extracted facts increases.}
  \label{tab:mine_frag_triples}
\end{table*}

\subsection{Implementation Details}
For all LLMs, we set the decoding temperature to $0$ to ensure reproducibility, except for GraphRAG, which follows the default stochastic configuration of its official implementation.

We adopted the canonicalization and factual retention evaluation protocol proposed by \citet{mo2025kggen}.
For the semantic clustering mentioned in Section~\ref{subsec:graph-construction-triples}, we partitioned entities and relations into clusters containing at most 128 elements.
For the identification of potential matches, we set the candidate retrieval size to $k=16$, which defines the number of top-ranked duplicates evaluated by the LLM.
All embedding-based processes used the all-MiniLM-L6-v2 model, and factual verification was performed via an LLM-as-a-judge protocol using GPT-4o.

For the downstream HotpotQA experiments, GPT-4.1 served as the answering model, and all-MiniLM-L6-v2 model was used for KG's triple retrieval and Naive RAG's chunk retrieval.

\section{Results and Discussion}

\subsection{Factual Retention Performance}
Table~\ref{tab:main_results} summarizes the factual retention performance on the MINE benchmark.
Across all compared methods and evaluated LLMs, SoKG consistently achieves the highest scores, peaking at 96.3\% with Claude-4.

The comparison between Direct Extraction and SoKG (w/o 5W1H) illustrates the benefit of introducing QA as an intermediate representation.
Even without 5W1H guidance, SoKG outperforms Direct Extraction on all LLM models.
This advantage stems from decomposing documents into discrete, self-contained QA pairs prior to triples extraction.

Notably, both GraphRAG and KGGen underperform Direct Extraction in terms of factual retention.
GraphRAG prioritizes hierarchical community structures and query-focused summarization over comprehensive fact preservation, resulting in lower coverage of atomic facts.
KGGen's entity-first bottleneck similarly leads to fact omission when initial entity identification fails, showing inconsistent performance across models.

This relative advantage of Direct Extraction reflects its lack of structural constraints: by avoiding early filtering or consolidation, it preserves a larger volume of raw triples.
However, as shown in Tables~\ref{tab:statistics} and \ref{tab:mine_frag_triples}, this comes at the cost of higher fragmentation, limiting the resulting graph's utility for downstream reasoning.

In contrast, SoKG with 5W1H guidance further enhances performance by systematically surfacing procedural and causal dimensions.
This interrogative framework ensures that latent dependencies are explicitly captured, maintaining high factual consistency regardless of the underlying LLM model's inherent reasoning capacity.

To further validate our triple extraction strategy, we conducted additional experiments in Appendix~\ref{sec:appendix_ablation}.
By isolating the impact of the QA scaffold from the extraction strategy, these studies reveal that entity-first approaches persist as a performance bottleneck even when applied to QA-preprocessed inputs.

\subsection{Graph Scale and Connectivity}
The superior factual retention shown in Table~\ref{tab:main_results} raises a critical question: is SoKG simply extracting more triples, or is it building a fundamentally better graph? 
To address this, we examine graph scale and connectivity in Table~\ref{tab:statistics}.

SoKG significantly expands graph scale while maintaining or improving connectivity density across all evaluated LLMs.
In contrast, Direct Extraction produces smaller graphs with lower connectivity, while GraphRAG generates compact structures that sacrifice comprehensive fact coverage for hierarchical organization.
The comparison between SoKG (w/o 5W1H) and SoKG reveals that 5W1H guidance substantially increases the number of extracted entities and relations while enhancing connectivity density.
This indicates that 5W1H systematically surfaces additional facts without fragmenting the graph structure.

Importantly, SoKG achieves higher connectivity than both Direct Extraction and KGGen despite using the same canonicalization procedure.
This confirms that the structural advantage originates from the QA-mediated intermediate representation, enabling relevant evidence to co-locate within 2-hop neighborhoods and directly supporting the high fact recoverability in Table~\ref{tab:main_results}.

Moreover, this increase in graph scale does not appear to reflect merely redundant expansion.
Supplementary analysis following recent quality-oriented evaluation perspectives for generative relation extraction~\citep{jiang2024genres} confirms that SoKG maintains competitive uniqueness, granularity, and factual precision even as the number of extracted triples substantially increases (Appendix~\ref{sec:appendix_quality}).

\subsection{Factual Volume and Structural Cohesion}
To further examine the relationship between knowledge coverage and graph fragmentation, we analyze triple counts and the NFI in Table~\ref{tab:mine_frag_triples}.
SoKG substantially expands knowledge volume while simultaneously reducing fragmentation across all evaluated LLMs.
While alternative methods either limit fact extraction (GraphRAG) or exhibit higher fragmentation (KGGen and Direct Extraction), SoKG extracts substantially more triples while maintaining lower NFI values.

The comparison between SoKG (w/o 5W1H) and SoKG (Ours) further illustrates the effectiveness of 5W1H guidance.
Adding 5W1H consistently increases triple extraction volume while reducing or maintaining similar fragmentation levels.
This pattern indicates that 5W1H not only surfaces additional facts but also enhances their integration into the graph structure.

\subsection{Downstream Multi-hop Reasoning} \label{subsec:downstream}
Table~\ref{tab:hotpotqa_main} reports multi-hop reasoning accuracy on HotpotQA Hard Bridge samples.
SoKG achieves the best overall performance across both backbones and retrieval depths.
Notably, while several graph-based methods such as KGGen and Direct Extraction fall below Naive RAG in certain settings, SoKG consistently outperforms it across all evaluated conditions, demonstrating that graph-structured retrieval is not inherently advantageous but becomes so when the graph is sufficiently connected and factually complete.

The largest gain is observed with Claude-4 under 3-hop retrieval, where SoKG reaches 56.38\%, outperforming Naive RAG by 8.5 percentage points and Direct Extraction by 16.5 percentage points.
GraphRAG also performs competitively under Claude-4, likely benefiting from its hierarchical community structure being better leveraged by stronger LLMs.
On Qwen-2.5, SoKG similarly leads at 27.00\% under 3-hop retrieval, outperforming Naive RAG by 6.9 percentage points.

A comparison between SoKG and SoKG (w/o 5W1H) further indicates that 5W1H-guided expansion improves not only factual retention but also the downstream usefulness of the resulting graph.
By explicitly externalizing latent connections as navigable edges, SoKG provides a more effective representation for tasks that require reasoning across disparate facts.

\begin{figure*}[t]
    \phantomsection
    \centering
    \includegraphics[width=0.99\textwidth]{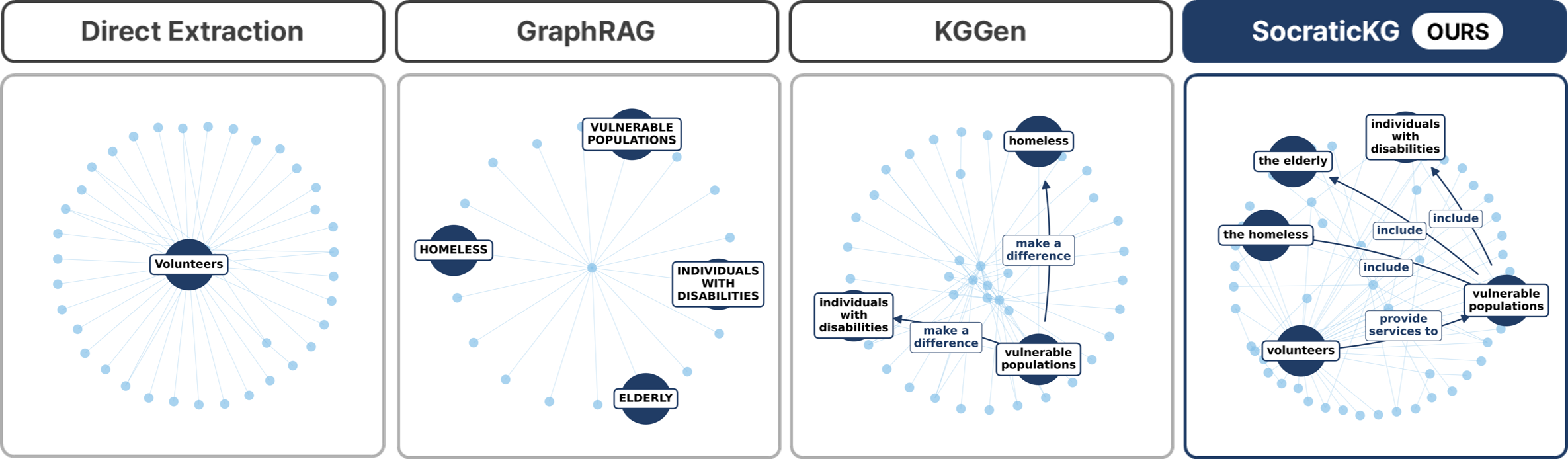}
    \caption{Comparison of extracted graphs for the example sentence: \textit{``Volunteers provide essential services and support to vulnerable populations, such as the homeless, the elderly, and individuals with disabilities.''} The nested relational path implied by this text (\textit{Volunteers} $\rightarrow$ \textit{Vulnerable Populations} $\rightarrow$ \{\textit{Homeless, Elderly, Individuals with disabilities}\}) is emphasized to assess relational completeness. Specifically, nodes corresponding to this path are enlarged for clear visibility, connected by thick dark blue arrows to indicate the sequence of triples, while the remaining background graph elements are displayed in light blue.}
    \label{fig:graph_viz}
\end{figure*}

\begin{table}[t]
  \centering
  \small
  \setlength{\tabcolsep}{0pt}
  \renewcommand{\arraystretch}{1.1}
  \begin{tabular*}{\columnwidth}{@{\extracolsep{\fill}} l cc cc}
    \toprule
    \multirow{2.3}{*}{\textbf{Method}}
      & \multicolumn{2}{c}{\textbf{Qwen-2.5}}
      & \multicolumn{2}{c}{\textbf{Claude-4}} \\
    \cmidrule(lr){2-3} \cmidrule(lr){4-5}
      & 2-hop & 3-hop & 2-hop & 3-hop \\
    \midrule
    Direct Ext.       & 19.50 & 23.88 & 37.87 & 39.88 \\
    GraphRAG          & 23.00 & 25.00 & 46.62 & 52.12 \\
    KGGen             & 16.50 & 18.50 & 38.25 & 46.75 \\
    SoKG (w/o 5W1H)   & 20.12 & 24.75 & 48.00 & 53.12 \\
    \textbf{SoKG}     & \textbf{23.62} & \textbf{27.00}
                      & \textbf{48.50} & \textbf{56.38} \\
    \midrule
    Naive RAG         & --    & 20.13 & --    & 47.88 \\
    \bottomrule
  \end{tabular*}
  \caption{Multi-hop reasoning accuracy (\%) on HotpotQA Hard Bridge samples. SoKG consistently performs best across backbones and retrieval depths, indicating the benefit of connected graph structure for multi-hop fact integration.}
  \label{tab:hotpotqa_main}
\end{table}

\subsection{Qualitative Analysis}
The cases in Figures~\ref{fig:case_study} and \ref{fig:graph_viz} provide concrete examples of how SoKG's interrogative process resolves the structural deficiencies and information loss observed in alternative methods.

Figure~\ref{fig:case_study} illustrates SoKG’s capacity to preserve logical coherence in complex participle phrases, such as \textit{facilitating cross-pollination}. 
GraphRAG, relying on entity summarization, fails to capture the causal structure entirely, while Direct Extraction and KGGen fragment or simplify the relationship.
In contrast, SoKG articulates the mediating concept to ensure a cohesive causal chain: \textit{bees} $\rightarrow$ \textit{cross-pollination} $\rightarrow$ \textit{genetic diversity}.

Similarly, Figure~\ref{fig:graph_viz} demonstrates how SoKG resolves relational fragmentation in nested entity structures.
GraphRAG isolates key entities as disconnected nodes, while KGGen introduces imprecise predicates such as \textit{make a difference}, losing the nested relational structure. SoKG fully reconstructs the relational tree by identifying all key entities and linking them via precise predicates such as \textit{provide services to} and \textit{include}.

These examples demonstrate that QA-mediated semantic scaffolding, guided by 5W1H inquiry, systematically addresses both causal reconstruction and relational completeness.
We further provide an qualitative analysis of failure cases in Appendix~\ref{sec:appendix_failure}, focusing on the QA mediation stage where the primary bottleneck lies.

\section{Conclusion}
We present SoKG, LLM-based KG construction method that uses QA pairs as a structured intermediate representation for document-level semantic expansion prior to triple extraction.
By employing 5W1H-guided QA generation, SoKG resolves referential ambiguities and surfaces implicit relational dependencies, ensuring that subsequent structural mapping is grounded in explicit, contextualized entities rather than underspecified inferences.

Evaluation on the MINE benchmark demonstrates that SoKG achieves superior factual coverage while simultaneously improving structural cohesion across diverse LLMs.
This performance stems from the QA-mediated scaffold, which systematically externalizes latent causal and relational dependencies that enhance graph connectivity even as the volume of extracted facts increases.

Furthermore, downstream experiments on HotpotQA confirm that these structural advantages translate into practical reasoning gains.
While other KG-based methods show inconsistent improvements over Naive RAG, SoKG outperforms it across all evaluated settings, demonstrating that QA-mediated construction produces KGs that are useful for complex multi-hop reasoning.

Our findings indicate that explicit semantic organization through QA generation is not merely an auxiliary preprocessing step but a key component for maintaining graph fidelity in LLM-based KG construction, enabling more structured knowledge extraction. 
By addressing the inherent trade-off between factual coverage and structural connectivity, SoKG provides a more reliable foundation for document-grounded knowledge representation and structured reasoning.

\clearpage
\section*{Limitations}
While SoKG's multi-stage pipeline naturally involves higher token consumption than single-pass extraction, we provide a detailed cost analysis in Appendix~\ref{sec:appendix_efficiency}.
Improving efficiency while maintaining factual density remains an important direction for future work.

Furthermore, as the graph quality depends on the reasoning depth of the underlying LLM, performance may vary in domains requiring highly specialized interrogative logic.
In particular, our failure analysis (Appendix~\ref{sec:appendix_failure}) reveals that the primary bottleneck lies in uni-dimensional queries that lack sufficient specificity, suggesting that more targeted questioning strategies could further enhance extraction quality.

Regarding graph representation, our current use of binary triples may simplify multidimensional qualifiers (\eg temporal or spatial data) that could be more compactly encoded via n-ary relations.
Finally, our evaluation focuses on factual recoverability and downstream multi-hop reasoning.
While this aligns with our objective of preserving document-level semantics, other dimensions---such as schema-alignment and relation-type fidelity---are left as promising avenues for the community to explore as KG evaluation standards evolve.

\section*{Ethical Considerations}
This study utilizes the publicly available MINE benchmark and LLMs.
We acknowledge that the benchmark and underlying LLMs may possess inherent biases, which could be reflected in the constructed graphs.
Additionally, automated extraction carries a risk of hallucinating facts not present in source documents.
We recommend human verification and validation for applications in sensitive or high-stakes domains.

\section*{Acknowledgements}
This work was supported by
the Technology Innovation Program funded by the Ministry of Trade, Industry and Energy (MOTIE, Korea) (RS-2025-25453780);
the National Research Foundation of Korea (RS-2023-00302123) funded by the Korean government, as part of the European Commission’s Horizon Europe framework programme (Grant Agreement No. 101135576, INTEND);
the Institute of Information and Communications Technology Planning and Evaluation (IITP) grant funded by the Korean government (MSIT) [RS-2021-II211343, Artificial Intelligence Graduate School Program (Seoul National University)];
and the Creative-Pioneering Researchers Program through Seoul National University.

\bibliography{custom}
\clearpage
\appendix

\twocolumn[
\section{Ablation Studies} 
\label{sec:appendix_ablation}

\begin{center}
\small
\setlength{\tabcolsep}{0pt}
\renewcommand{\arraystretch}{1.1}
  \begin{tabular*}{\textwidth}{@{\extracolsep{\fill}} l c c c c c c c c c c }
    \toprule
    \multirow{2.3}{*}{\textbf{Prompt Archetype}}
      & \multicolumn{2}{c}{\textbf{Qwen-2.5}} & \multicolumn{2}{c}{\textbf{GPT-4o-mini}} & \multicolumn{2}{c}{\textbf{GPT-4o}} & \multicolumn{2}{c}{\textbf{Gemini-2.5}} & \multicolumn{2}{c}{\textbf{Claude-4}} \\
    \cmidrule(lr){2-3} \cmidrule(lr){4-5} \cmidrule(lr){6-7} \cmidrule(lr){8-9} \cmidrule(lr){10-11}
      & w/o 5W1H & Full & w/o 5W1H & Full & w/o 5W1H & Full & w/o 5W1H & Full & w/o 5W1H & Full \\
    \midrule
    Role-Oriented (RO) & 67.1 & \textbf{73.4} & 80.5 & \textbf{83.9} & 83.5 & \textbf{89.3} & 85.6 & \textbf{87.7} & 94.6 & \textbf{96.3} \\
    Procedural-Step (PS) & 60.7 & \textbf{64.5} & 79.6 & \textbf{83.0} & 82.1 & \textbf{87.1} & 86.7 & \textbf{87.3} & 93.5 & \textbf{95.8} \\
    Instructional-Direct (ID) & 65.3 & \textbf{68.6} & 77.9 & \textbf{83.5} & 79.7 & \textbf{81.2} & 83.9 & \textbf{88.7} & 91.3 & \textbf{96.3} \\
    \midrule
    \textbf{Average} & 64.4 & \textbf{68.8} & 79.3 & \textbf{83.5} & 81.8 & \textbf{85.9} & 85.4 & \textbf{87.9} & 93.1 & \textbf{96.1} \\
    \bottomrule
  \end{tabular*}
  \captionof{table}{Effectiveness of 5W1H-guided expansion across prompt archetypes. Scores represent factual retention (\%) on the MINE benchmark. The results demonstrate that the 5W1H framework is a robust cognitive guide independent of stylistic framing.}
  \label{tab:prompt_ablation}
\end{center}

\vspace{0.5em}

\begin{center}
\small
\setlength{\tabcolsep}{0pt}
\renewcommand{\arraystretch}{1.1}

\begin{tabular*}{\textwidth}{@{\extracolsep{\fill}} l c c c c c }
\toprule
\textbf{Method} & \textbf{Qwen-2.5} & \textbf{GPT-4o-mini} & \textbf{GPT-4o} & \textbf{Gemini-2.5} & \textbf{Claude-4} \\
\midrule
KGGen          & 56.7 & 44.3 & 66.4 & 62.5 & 69.1 \\
SoKG-EF        & 72.1 & 79.7 & 76.9 & 87.0 & 92.5 \\
\textbf{SoKG (Ours)} & \textbf{73.4} & \textbf{83.9} & \textbf{89.3} & \textbf{87.7} & \textbf{96.3} \\
\bottomrule
\end{tabular*}

\captionof{table}{Ablation study on input representation and triple extraction strategy. While SoKG-EF demonstrates the foundational impact of the QA scaffold, SoKG achieves peak performance by removing the entity-first bottleneck to maximize factual retention across all LLMs.}
\label{tab:ablation}
\end{center}
\vspace{1em}
]

\subsection{Robustness of Prompt Designs}
\label{sec:appendix_prompt_ablation}
To evaluate the contribution of the 5W1H framework, Table~\ref{tab:prompt_ablation} compares the full 5W1H-integrated pipeline (Full) against the version omitting 5W1H-guided expansion (w/o 5W1H) across three distinct prompt archetypes:

\begin{itemize}
\item \textbf{Role-Oriented (RO):} Assigns a specific persona (e.g., Knowledge Archivist) and uses 5W1H as analytical lenses to guide deep exploration.
This prompt design was adopted as the primary setting for our main experiments.
\item \textbf{Procedural-Step (PS):} Defines a systematic workflow (Read $\rightarrow$ Segment $\rightarrow$ Generate) to ensure atomic factual extraction.
\item \textbf{Instructional-Direct (ID):} Employs standard task-based instructions without complex role-play or multi-step procedures.
\end{itemize}

As shown in Table~\ref{tab:prompt_ablation}, the 5W1H framework provides a universal performance lift across all LLMs regardless of the underlying prompt structure.
While the RO archetype generally yields the highest retention, peaking at $96.3\%$ with Claude-4, even the more concise PS and ID templates show significant improvements once the interrogative scaffold is present. 
These results indicate that the 5W1H constraint functions as a fundamental cognitive guide that systematically surfaces procedural and causal dimensions.

\subsection{The Entity-First Constraint}
\label{sec:appendix_extraction_ablation}
We evaluate the respective impacts of the QA scaffold and extraction strategy by comparing three configurations: KGGen, SoKG-EF (Entity-First), and SoKG.
SoKG-EF incorporates both the extraction and consolidation logic of KGGen, applying this entity-centric pipeline to our QA-mediated scaffold.
This setup allows us to evaluate the benefit of the scaffold independently while preserving the underlying entity-first logic.

As shown in Table~\ref{tab:ablation}, the superior performance of SoKG-EF over KGGen confirms that a QA scaffold effectively mitigates the complexity of raw text. 
However, SoKG’s even greater success reveals that rigid entity-first filtering acts as a restrictive bottleneck, limiting the model's ability to capture full relational depth even when supported by a comprehensive QA scaffold.

Although intermediate stages introduce overhead, this rich interrogative structure justifies the investment by providing a dense semantic foundation for superior factual recoverability. 
Unlike isolated entity extraction which often incurs information loss through restrictive filtering, the QA-mediated scaffold preserves a richer semantic context. 
These results demonstrate that for structural refinement, a QA-driven approach offers a more systematic and inclusive foundation for knowledge construction than traditional entity-centric methods.
 
\onecolumn
\section{QA Generation Prompt Details}
\label{sec:appendix_prompts}

\subsection{Role-Oriented (RO), w/ 5W1H}
\label{sec:5w1h_prompts}
\begin{lstlisting}
## ROLE
You are a **Comprehensive Knowledge Archivist** who converts the [Full Document] into detailed, document-grounded QA pairs.

## OBJECTIVE
Extract as many meaningful Question-Answer pairs as possible from the document.
Use the 5W1H perspectives (Who, What, When, Where, Why, How) **as analytical lenses** to help you identify and expand potential questions, but do NOT restrict yourself to producing only 5W1H-type questions.
Your goal is to maximize informational coverage, capturing every explicit fact, relation, event, definition, rationale, and process described in the document.

## INPUT
Full Document: "{document_text}"

## CONSTRAINTS
1. **Context-Independent**
   - Each QA must be self-contained and understandable without referencing the original text.
   - Replace pronouns with explicit entities.

2. **No Hallucination**
   - Use only facts explicitly stated in the document.

3. **Expansion-Oriented Thinking**
   - For each sentence or factual unit, consider the 5W1H perspectives as prompts to explore:
     - WHO is involved?
     - WHAT happened or is described?
     - WHEN did it occur?
     - WHERE did it occur?
     - WHY did it occur?
     - HOW was it carried out?
   - These perspectives are **guides** to inspire multiple possible QA pairs, even if they are implicit or only partially expressed.

4. **Coverage**
   - Extract all possible QA pairs that can be reasonably derived from the document.

## OUTPUT FORMAT
Return a JSON list of QA objects:

[
  {{"question": "...", "answer": "..."}},
  ...
]
\end{lstlisting}
\clearpage
\subsection{Role-Oriented (RO), w/o 5W1H}
\begin{lstlisting}
## ROLE
You are a **Comprehensive Knowledge Archivist** who converts the [Full Document] into precise and meaningful QA pairs.

## OBJECTIVE
Extract as many high-quality Question-Answer pairs as needed to fully represent the document's explicit information.
Use the following analytical perspectives as guides to discover potential questions, but do NOT restrict your output to only these categories:

1. **Entities & Definitions** - Identify and clarify key terms, objects, roles, or concepts.
2. **Properties & Characteristics** - Extract notable features, attributes, components, or qualities.
3. **Events & Stated Facts** - Capture actions, processes, or explicit factual statements.
4. **Relationships & Dependencies** - Identify connections, comparisons, or dependencies between entities or ideas.

These perspectives are **guides for expanding coverage**, not mandatory categories.

## INPUT
Full Document: "{document_text}"

## CONSTRAINTS
1. **Context-Independent**
   - Each QA must be self-contained and understandable without referencing the original text.
   - Replace pronouns with explicit entities when needed.

2. **No Hallucination**
   - Use only facts explicitly stated in the document.

3. **Coverage without Inflation**
   - Extract all meaningful QA pairs that can be reasonably derived from the document.

## OUTPUT FORMAT
Return a JSON list:

[
  {{"question": "...", "answer": "..."}},
  ...
]
\end{lstlisting}
\clearpage
\subsection{Procedural-Step (PS), w/ 5W1H}
\begin{lstlisting}
## ROLE
You are a **Document-Grounded QA Extractor**.

## OBJECTIVE
Convert the full document into high-coverage, explicit-fact QA pairs.

## PROCEDURE
1. Read the document end-to-end.
2. Segment into atomic factual units.
3. For each unit:
   - Generate QAs that capture all explicit information it contains.
   - When forming questions, view the unit through the 5W1H angles (Who, What, When, Where, Why, How) so that different aspects of the same fact can be covered.
4. Merge duplicates and keep the most precise wording.

## INPUT
Full Document: "{document_text}"

## CONSTRAINTS
- Context-Independent QAs only.
- No Hallucination.
- Prefer concise but complete answers.

## OUTPUT FORMAT
Return a JSON list:
[
  {{"question": "...", "answer": "..."}},
  ...
]
\end{lstlisting}
\subsection{Procedural-Step (PS), w/o 5W1H}
\begin{lstlisting}
## ROLE
You are a **Document-Grounded QA Extractor**.

## OBJECTIVE
Convert the full document into high-coverage, explicit-fact QA pairs.

## PROCEDURE
1. Read the document end-to-end.
2. Segment into atomic factual units.
3. For each unit, generate QAs that capture all explicit information it contains.
4. Merge duplicates and keep the most precise wording.

## INPUT
Full Document: "{document_text}"

## CONSTRAINTS
- Context-Independent QAs only.
- No Hallucination.
- Prefer concise but complete answers.

## OUTPUT FORMAT
Return a JSON list:
[
  {{"question": "...", "answer": "..."}},
  ...
]
\end{lstlisting}
\clearpage
\subsection{Instructional-Direct (ID), w/ 5W1H}
\begin{lstlisting}
Read the following document and generate question-answer pairs based on its content.
Generate as many high-quality questions as needed to cover the information explicitly stated in the document.
For the same piece of information, consider the 5W1H dimensions (Who, What, When, Where, Why, How) and generate separate questions whenever different aspects are supported by the text.
Do not stop at a single question if multiple 5W1H aspects can be identified.
If different parts of the document support different questions, include all of them.
Each question should be answerable using information explicitly stated in the document and written in a clear and self-contained manner.

Input Document:
"{document_text}"

Output Format:
Return a JSON list of objects in the following form:
[
  {{"question": "...", "answer": "..."}},
  ...
]
\end{lstlisting}
\subsection{Instructional-Direct (ID), w/o 5W1H}
\begin{lstlisting}
Read the following document and generate question-answer pairs based on its content.
Generate as many high-quality questions as needed to cover the information explicitly stated in the document.
If different parts of the document support different questions, include all of them.
Each question should be answerable using information explicitly stated in the document and written in a clear and self-contained manner.

Input Document:
"{document_text}"

Output Format:
Return a JSON list of objects in the following form:
[
  {{"question": "...", "answer": "..."}},
  ...
]
\end{lstlisting}
\clearpage
\section{Triple Extraction Prompt Details}
\subsection{Triple Extraction from QA Pairs}
\label{sec:triple_prompts}
\begin{lstlisting}
## ROLE
You are a Semantic Knowledge Graph Builder.
Extract every structured triples (entity1, relation, entity2) from the Q&A pair, following the rules below.

## GOAL
From the question-answer pair, extract only useful, knowledge-ready triples that can serve as entries in a semantic knowledge graph.

## RULES
Extract clean (subject, relation, object) triples following the rules:

1. Split every stated or clearly implied fact into minimal triples; integrate question and answer context when needed.

2. Entities (entity1, entity2) must be short, concrete noun phrases.
   - No pronouns (this, that, it, its, these, those, etc.).
   - Entities must not be unresolved or reference-based pronouns (\eg those, they, someone, anyone, whoever); if such a pronoun appears, rewrite it into a specific, explicit noun phrase or skip the triple.
   - No clauses or relative clauses (no "who/that/which/what/as it ..." inside an entity).
   - No long gerund or sentence-like phrases. If a phrase contains a verb or clause marker, rewrite it into a concise noun concept or skip the triple.

3. Relations must be short, canonical verbs or verb phrases.
   - Express a single semantic link between the two entities (\eg causes, leads to, supports, believes, opposes).
   - Must be a compact predicate, not a sentence fragment.
   - No pronouns or clause markers inside the relation (no "its", "that", "as it", "what", etc.).
   - If the source uses an idiomatic or long expression, rewrite it into a simple canonical relation without pronouns or embedded clauses, or skip the triple.

4. Include a fact if it can be clearly rewritten into a concise, explicit triple that fits the rules above; otherwise skip it.

5. Output only concise, interpretable, knowledge-ready triples.

## INPUT
Q: {question}
A: {answer}

## OUTPUT FORMAT (JSON List)
- Return a list of JSON objects.
- Return [] if no valid triples exist.

[
  {{"entity1": "Specific_Noun", "relation": "precise_verb_phrase", "entity2": "Specific_Noun"}}
]
\end{lstlisting}
\clearpage
\subsection{Triple Extraction from Raw Text (Direct Extraction)}
\label{sec:direct_extraction_prompt}
\begin{lstlisting}
## ROLE
You are a Semantic Knowledge Graph Builder.
Extract every structured triples (entity1, relation, entity2) from the text, following the rules below.

## GOAL
From the given text, extract only useful, knowledge-ready triples that can serve as entries in a semantic knowledge graph.

## RULES
Extract clean (subject, relation, object) triples following the rules:

1. Split every stated or clearly implied fact into minimal triples.

2. Entities (entity1, entity2) must be short, concrete noun phrases.
   - No pronouns (this, that, it, its, these, those, etc.).
   - Entities must not be unresolved or reference-based pronouns (\eg those, they, someone, anyone, whoever); if such a pronoun appears, rewrite it into a specific, explicit noun phrase or skip the triple.
   - No clauses or relative clauses (no "who/that/which/what/as it ..." inside an entity).
   - No long gerund or sentence-like phrases. If a phrase contains a verb or clause marker, rewrite it into a concise noun concept or skip the triple.

3. Relations must be short, canonical verbs or verb phrases.
   - Express a single semantic link between the two entities (\eg causes, leads to, supports, believes, opposes).
   - Must be a compact predicate, not a sentence fragment.
   - No pronouns or clause markers inside the relation (no "its", "that", "as it", "what", etc.).
   - If the source uses an idiomatic or long expression, rewrite it into a simple canonical relation without pronouns or embedded clauses, or skip the triple.

4. Include a fact if it can be clearly rewritten into a concise, explicit triple that fits the rules above; otherwise skip it.

5. Output only concise, interpretable, knowledge-ready triples.

## INPUT
Text: {document_text}

## OUTPUT FORMAT (JSON List)
- Return a list of JSON objects.
- Return [] if no valid triples exist.

[
  {{"entity1": "Specific_Noun", "relation": "precise_verb_phrase", "entity2": "Specific_Noun"}}
]
\end{lstlisting}

\twocolumn[
\section{Computational Efficiency Analysis}
\label{sec:appendix_efficiency}

\begin{center}
\small
\setlength{\tabcolsep}{0pt}
\renewcommand{\arraystretch}{1.1}
\begin{tabular*}{\textwidth}{@{\extracolsep{\fill}} l rrc rrc}
  \toprule
  \multirow{2}{*}{\textbf{Method}}
    & \multicolumn{3}{c}{\textbf{Qwen-2.5}}
    & \multicolumn{3}{c}{\textbf{Claude-4}} \\
  \cmidrule(lr){2-4} \cmidrule(lr){5-7}
    & $T_{\text{total}}$ & $N_{\text{tri}}$ & $\tau$
    & $T_{\text{total}}$ & $N_{\text{tri}}$ & $\tau$ \\
  \midrule
  Direct Extraction
    & 145,262 & 1,955 & 74.30
    & 150,569 & 4,417 & 34.09 \\
  GraphRAG
    & 333,646 & 1,981 & 168.42
    & 331,313 & 1,590 & 208.37 \\
  KGGen
    & 231,762 & 2,375 & 97.58
    & 256,691 & 6,391 & 40.16 \\
  SoKG (w/o 5W1H)
    & 275,137 & 2,871 & 95.83
    & 392,284 & 10,511 & 37.32 \\
  \textbf{SoKG (Ours)}
    & 333,917 & 3,958 & 84.36
    & 553,530 & 14,849 & 37.28 \\
  \bottomrule
\end{tabular*}

\captionof{table}{Computational cost analysis on the MINE benchmark for Qwen-2.5 (weakest backbone) and Claude-4 (strongest backbone). $T_{\text{total}}$: total tokens consumed up to triple extraction (summed over 100 articles); $N_{\text{tri}}$: canonicalized triple count; $\tau = T_{\text{total}} / N_{\text{tri}}$: tokens per triple.}
\label{tab:efficiency_full}
\end{center}
]

The multi-stage QA pipeline naturally incurs higher token consumption than single-pass extraction.
To quantify this overhead, Table~\ref{tab:efficiency_full} reports the total token consumption (input and output) accumulated up to the triple extraction stage for each method, along with the resulting triple count after canonicalization.

SoKG consumes approximately 2--3$\times$ more total tokens than Direct Extraction across both backbones, reflecting the additional cost of QA generation and per-pair triple extraction.
However, a substantial portion of this investment is converted into a larger volume of canonicalized triples: SoKG produces 2.0$\times$ more triples than Direct Extraction on Qwen-2.5 and 3.4$\times$ more on Claude-4.
 
Direct Extraction achieves the lowest absolute token consumption, but as shown in the main results (Tables~\ref{tab:main_results}--\ref{tab:mine_frag_triples}), the resulting graphs exhibit higher fragmentation and lower factual retention, limiting their downstream utility.
GraphRAG and KGGen consume comparable or higher total tokens than SoKG on certain backbones while producing substantially fewer triples, resulting in less favorable cost-to-knowledge ratios.
 
These results indicate that while the QA-mediated pipeline introduces meaningful computational overhead, the additional cost is offset by a proportionally larger gain in extracted knowledge volume and structural quality.
Reducing this overhead through selective QA generation or adaptive questioning depth remains a promising direction for future work.

\section{Triple Quality and Hallucination Analysis}
\label{sec:appendix_quality}

To verify that SoKG's expanded extraction volume does not introduce redundancy or hallucination, we evaluated triple quality using two automated metrics from the GenRES framework~\citep{jiang2024genres}---Uniqueness Score (US) and Granularity Score (GS)---alongside human-evaluated Factual Precision (FP).
We conduct this analysis on triples extracted from the MINE benchmark using Qwen2.5-7B-Instruct, the lightest backbone in our evaluation and thus most susceptible to quality degradation.

\paragraph{Uniqueness Score (US)}
measures pairwise diversity among extracted triples within each document.
For each triple $t_i$, we compute its maximum cosine similarity to all other triples in the same document and define uniqueness as $\text{US}(t_i) = 1 - \max_{j \neq i} \cos(t_i, t_j)$.
The document-level US is then averaged over all triples.

\paragraph{Granularity Score (GS)}
evaluates the atomicity of each triple by penalizing overly broad facts that could be further decomposed.
Following \citet{jiang2024genres}, an LLM judge assesses whether each triple represents a single, indivisible fact.

\paragraph{Factual Precision (FP)}
was assessed through manual verification.
Five human annotators independently evaluated all triples extracted from three randomly sampled MINE document against the source text, labeling each triple as factually supported or not.

\begin{table}[!htbp]
  \centering
  \small
  \setlength{\tabcolsep}{0pt}
  \renewcommand{\arraystretch}{1.1}
  \begin{tabular*}{\columnwidth}{@{\extracolsep{\fill}} l r c c c}
    \toprule
    \textbf{Method} & $N_{\text{tri}}$ & US (\%) & GS & FP (\%) \\
    \midrule
    Direct Ext.
      & 1,955 & 88.02 & 69.09
      & 86.42{\scriptsize $\pm$4.21} \\
    GraphRAG
      & 1,981 & 86.86 & 54.24
      & 83.15{\scriptsize $\pm$7.38} \\
    KGGen
      & 2,375 & 83.66 & 87.72
      & 83.93{\scriptsize $\pm$8.67} \\
    SoKG (w/o 5W1H)
      & 2,871 & 90.62 & 82.03
      & 75.12{\scriptsize $\pm$9.45} \\
    SoKG
      & 3,958 & 91.75 & 82.59
      & 84.21{\scriptsize $\pm$6.33} \\
    \bottomrule
  \end{tabular*}
  \caption{Triple quality analysis on Qwen-2.5. US: Uniqueness Score (\%, higher is better); GS: Granularity Score (higher is better); FP: human-evaluated Factual Precision (\%).}
  \label{tab:quality}
\end{table}

As shown in Table~\ref{tab:quality}, SoKG achieves the highest Uniqueness Score (91.75\%), confirming that 5W1H-guided expansion surfaces distinct, non-repetitive information even as extraction density increases.
The Granularity Score (82.59) substantially exceeds those of Direct Extraction (69.09) and GraphRAG (54.24), indicating that the QA scaffold enforces atomic precision rather than producing overly broad triples.

For Factual Precision, SoKG achieves 84.38\%---comparable to Direct Extraction (88.00\%) and consolidation-based methods.
Direct Extraction's slightly higher precision reflects its tendency to capture only surface-level, explicit facts that are inherently easier to verify.
Notably, removing the 5W1H scaffold (SoKG w/o 5W1H) reduces precision to 74.55\%, suggesting that systematic interrogative guidance is critical for maintaining grounded construction as extraction volume increases.

These results collectively demonstrate that SoKG's expanded knowledge volume reflects genuine informational gain rather than redundant or hallucinated content.

\section{Failure Analysis of QA Mediation}
\label{sec:appendix_failure}

To provide a transparent account of the framework's limitations, we analyzed failure patterns in the QA mediation stage---the primary contribution of this work---rather than triple extraction, which modern LLMs handle with relative proficiency.
Our analysis revealed minimal factual hallucinations during QA generation; the generated pairs were largely grounded in the source text.
The primary failure mode was not a lack of truth but a lack of \textit{semantic utility}: the QA pairs failed to surface sufficiently specific or discriminative information for downstream triple extraction.

These failures stem almost exclusively from uni-dimensional ``What'' queries that exhibit the following recurring patterns:

\paragraph{Lack of Specificity}
The question targets a broad topic rather than a specific mechanism, yielding answers that list entities without clarifying their functional roles.

\begin{quote}
\small
\textbf{Q:} What new challenges have emerged with recent military technology advancements? \\
\textbf{A:} The rise of UAVs, drones, and AI has introduced new challenges and opportunities.
\end{quote}

\noindent
The answer enumerates technological entities rather than addressing the nature of the challenges themselves (\eg precision strikes, real-time surveillance), failing to capture the causal mechanism the question implies.

\paragraph{Sophisticated Non-Answers}
The answer restates the document's framing rather than grounding the response in specific evidence.

\begin{quote}
\small
\textbf{Q:} What is the essay exploring regarding military technology? \\
\textbf{A:} This essay will explore the evolution of military technology, highlighting key innovations.
\end{quote}

\noindent
The response remains at the level of the document's introduction, relying on general nouns (\eg \textit{technology}, \textit{innovation}) rather than extracting concrete facts from the body of the text.

\paragraph{Context Loss}
The answer is factually accurate but omits the specific contextual details that would produce discriminative triples.

\begin{quote}
\small
\textbf{Q:} What risk is associated with gene manipulation regarding individual rights? \\
\textbf{A:} There is a risk of infringing on individuals' rights to make informed decisions about their own genetic information.
\end{quote}

\noindent
While correct, the answer misses the specific examples provided in the source text---such as discrimination based on genetic traits---producing triples with low discriminative value for graph-based reasoning.
These failure modes highlight a fundamental limitation: the quality of QA mediation is bounded by the LLM's ability to formulate deep, specific queries.
The concentration of failures in ``What'' questions suggests that this category, while the most frequently generated, is also the most prone to shallow formulation.
More targeted prompt refinement or adaptive questioning strategies that dynamically adjust question depth based on document complexity represent promising directions for future work.

\end{document}